\documentclass{article}

\title{Naive Bayes and Text Classification I \\ \vspace{2 mm} \large Introduction and Theory}

\author{Sebastian Raschka\\ \texttt{se.raschka@gmail.com}} 
\date{October 4, 2014}
\usepackage{graphicx}
\usepackage{hyperref}
\usepackage{cite}
\usepackage{amsmath}
\usepackage[font=small,labelfont=bf]{caption}
\usepackage{float}

\usepackage{fancyhdr}
\begin{document} % start main document

\maketitle % makes the title page from meta data
\tableofcontents

\section{Introduction}
\label{sec:introduction}

Starting more than half a century ago, scientists became very serious about addressing the question: "Can we build a model that learns from available data and automatically makes the right decisions and predictions?" Looking back, this sounds almost like a rhetoric question, and the answer can be found in numerous applications that are emerging from the fields of pattern classification, machine learning, and artificial intelligence.  

Data from various sensoring devices combined with powerful learning algorithms and domain knowledge led to many great inventions that we now take for granted in our everyday life: Internet queries via search engines like Google, text recognition at the post office, barcode scanners at the supermarket, the diagnosis of diseases, speech recognition by Siri or Google Now on our mobile phone, just to name a few.

One of the sub-fields of \emph{predictive modeling} is \emph{supervised pattern classification}; supervised pattern classification is the task of training a model based on labeled training data which then can be used to assign a pre-defined class label to new objects. One example that we will explore throughout this article is spam filtering via naive Bayes classifiers in order to predict whether a new text message can be categorized as spam or not-spam.
Naive Bayes classifiers, a family of classifiers that are based on the popular Bayes' probability theorem, are known for creating simple yet well performing models, especially in the fields of document classification and disease prediction.

\begin{figure}[h!]
\includegraphics[width=\linewidth]{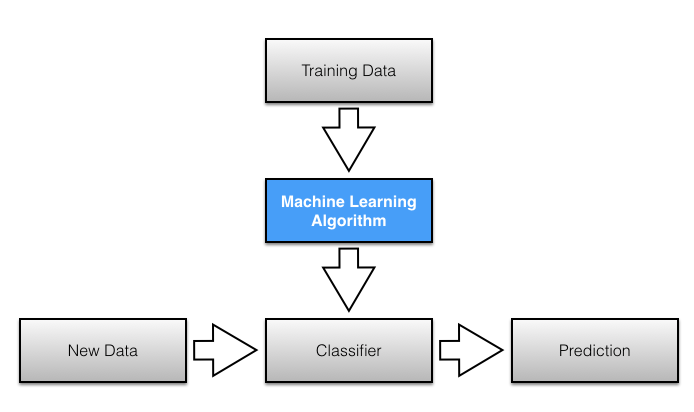}
\caption{A simplified diagram of the general model building procedure for pattern classification.}
\label{fig:learning_model}
\end{figure}

A more detailed overview of predictive modeling can be found in my previous article \emph{\href{http://sebastianraschka.com/Articles/2014_intro_supervised_learning.html}{Predictive Modeling, Supervised Machine Learning, and Pattern Classification - The Big Picture}}.

\section{Naive Bayes Classification}
\label{sec:naive_bayes_classification}

\subsection{Overview}
\label{sec:overview}

\emph{Naive Bayes classifiers} are linear classifiers that are known for being simple yet very efficient. The probabilistic model of naive Bayes classifiers is based on Bayes' theorem, and the adjective \emph{naive} comes from the assumption that the features in a dataset are mutually independent. In practice, the independence assumption is often violated, but naive Bayes classifiers still tend to perform very well under this unrealistic assumption \cite{rish2001empirical}. Especially for small sample sizes, naive Bayes classifiers can outperform the more powerful alternatives \cite{domingos1997optimality}.

Being relatively robust, easy to implement, fast, and accurate, naive Bayes classifiers are used in many different fields. Some examples include the diagnosis of diseases and making decisions about treatment processes \cite{kazmierska2008application}, the classification of RNA sequences in taxonomic studies \cite{wang2007naive}, and spam filtering in e-mail clients \cite{sahami1998bayesian}.  
However, strong violations of the independence assumptions and non-linear classification problems can lead to very poor performances of naive Bayes classifiers.  
We have to keep in mind that the type of data and the type problem to be solved dictate which classification model we want to choose. In practice, it is always recommended to compare different classification models on the particular dataset and consider the prediction performances as well as computational efficiency. 

In the following sections, we will take a closer look at the probability model of the naive Bayes classifier and apply the concept to a simple toy problem. Later, we will use a publicly available SMS (text message) collection to train a naive Bayes classifier in Python that allows us to classify unseen messages as spam or ham.

\begin{figure}[h!]
\includegraphics[width=\linewidth]{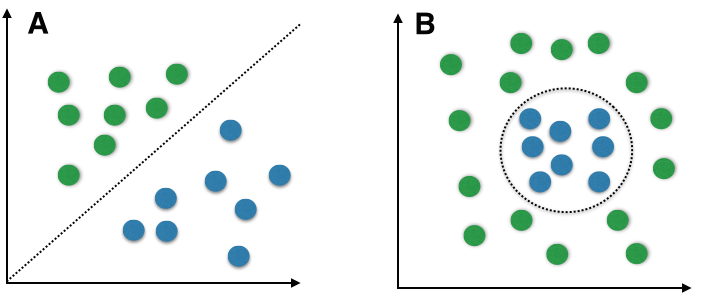}
\caption{Linear (A) vs. non-linear problems (B). Random samples for two different classes are shown as colored spheres, and the dotted lines indicate the class boundaries that classifiers try to approximate by computing the decision boundaries.
A non-linear problem (B) would be a case where linear classifiers, such as naive Bayes, would not be suitable since the classes are not linearly separable. In such a scenario, non-linear classifiers (e.g.,instance-based nearest neighbor classifiers) should be preferred.}
\label{fig:nonlinear_probs}
\end{figure}

\subsection{Posterior Probabilities}
\label{sec:posterior_probabilities_1}

In order to understand how naive Bayes classifiers work, we have to briefly recapitulate the concept of Bayes' rule.
The probability model that was formulated by Thomas Bayes (1701-1761) is quite simple yet powerful; it can be written down in simple words as follows:

\begin{equation} \text{posterior probability} = \frac{\text{conditional probability} \cdot \text{prior probability}}{\text{evidence}} \end{equation}

 Bayes' theorem forms the core of the whole concept of naive Bayes classification. The \emph{posterior probability}, in the context of a classification problem, can be interpreted as: "What is the probability that a particular object belongs to class $i$ given its observed feature values?" A more concrete example would be: "What is the probability that a person has diabetes given a certain value for a pre-breakfast blood glucose measurement and a certain value for a post-breakfast blood glucose measurement?"

\begin{equation}  P(\text{diabetes} \mid \textbf x_i) \;, \quad \textbf x_i = [90 \text{mg/dl}, 145 \text{mg/dl}] \end{equation}

 Let

\begin{itemize}
	\item $\textbf x_i$ be the feature vector of sample $i, \; i \in \{1, 2, ..., n\}$,
	\item $\omega_j$ be the notation of class $j, \; j \in \{1, 2, ..., m\}$, 
	\item and $P(\textbf x_i \mid \omega_j)$ be the probability of observing sample $\textbf x_i$ given that is belongs to class $\omega_j$.
\end{itemize}

 The general notation of the posterior probability can be written as

\begin{equation} P(\omega_j \mid \textbf x_i) = \frac{P(\textbf x_i \mid \omega_j) \cdot P(\omega_j)}{P(\textbf x_i)} \end{equation}

The objective function in the naive Bayes probability is to maximize the posterior probability given the training data in order to formulate the decision rule. 

\begin{equation} \text{predicted class label} \leftarrow \underset{j = 1 ..., m} {\text{arg max}} \; P(\omega_{j} \mid \textbf x_i)   \end{equation}

To continue with our example above, we can formulate the decision rule based on the posterior probabilities as follows:

\begin{equation} 
\begin{split} 
   & \text{person has diabetes if} \\
   & P(\text{diabetes} \mid \textbf x_i) \ge P(\text{not-diabetes} \mid \textbf x_i), \\
   & \text{else classify person as healthy}.
\end{split}
   \end{equation}

\subsection{Class-conditional Probabilities}
\label{sec:class-conditional_probabilities_1}

One assumption that Bayes classifiers make is that the samples are \emph{i.i.d.}  
The abbreviation \emph{i.i.d.} stands for "independent and identically distributed" and describes random variables that are independent from one another and are drawn from a similar probability distribution. Independence means that the probability of one observation does not affect the probability of another observation (e.g., time series and network graphs are not independent).  One popular example of  \emph{i.i.d.} variables is the classic coin tossing: The first coin flip does not affect the outcome of a second coin flip and so forth. Given a fair coin, the probability of the coin landing on "heads" is always 0.5 no matter of how often the coin if flipped.

An additional assumption of naive Bayes classifiers is the \emph{conditional independence} of features. Under this \emph{naive} assumption, the \emph{class-conditional probabilities} or (\emph{likelihoods}) of the samples can be directly estimated from the training data instead of evaluating all possibilities of \textbf{x}. Thus, given a $d$-dimensional feature vector \textbf{x}, the class conditional probability can be calculated as follows:

\begin{equation} P(\textbf x \mid \omega_j) = P(x_1 \mid \omega_j) \cdot P(x_2 \mid \omega_j) \cdot \ldots \cdot P(x_d \mid \omega_j) =  \prod_{k=1}^{d} P( x_k \mid \omega_j) \end{equation}

Here, $P(\textbf x \mid \omega_j)$ simply means: "How likely is it to observe this particular pattern $\textbf x$ given that it belongs to class $ \omega_j$?" The "individual" likelihoods for every feature in the feature vector can be estimated via the maximum-likelihood estimate, which is simply a frequency in the case of categorical data:
    
\begin{equation} \hat{P}(x_i \mid \omega_j) = \frac{N_{x_i, \omega_j}}{N_{\omega_j}}  \quad (i = (1, ..., d))\end{equation}

\begin{itemize}
	\item $N_{x_i, \omega_j}$: Number of times feature $x_i$ appears in samples from class $\omega_j$.
	\item  $N_{\omega_j}$: Total count of all features in class $\omega_j$.
\end{itemize}

To illustrate this concept with an example, let's assume that we have a collection of 500 documents where 100 documents are \emph{spam} messages. Now, we want to calculate the class-conditional probability for a new message "Hello World" given that it is spam.
Here, the pattern consists of two features: "hello" and "world," and the class-conditional probability is the product of the "probability of encountering 'hello' given the message is spam" --- the probability of encountering "world" given the message is spam."

\begin{equation} P(\textbf x=[\text{hello, world}] \mid \omega=\text{spam}) = P(\text{hello} \mid \text{spam}) \cdot P(\text{world} \mid \text{spam}) \end{equation}

Using the training dataset of 500 documents, we can use the maximum-likelihood estimate to estimate those probabilities: We'd simply calculate how often the words occur in the corpus of all spam messages. E.g.,

\begin{equation} \hat{P}(\textbf x=[\text{hello, world}] \mid \omega=\text{spam}) = \frac{20}{100} \cdot \frac{2}{100} = 0.004 \end{equation}

However, with respect to the \emph{naive} assumption of conditional independence, we notice a problem here: The \emph{naive} assumption is that a particular word does not influence the chance of encountering other words in the same document. For example, given the two words  "peanut" and "butter" in a text document, intuition tells us that this assumption is obviously violated: If a document contains the word "peanut" it will be more likely that it also contains the word "butter" (or "allergy"). In practice, the conditional independence assumption is indeed often violated, but naive Bayes classifiers are known to perform still well in those cases \cite{zhang2004optimality}.

\subsection{Prior Probabilities}
\label{sec:prior_probabilities_1}

In contrast to a frequentist's approach, an additional \emph{prior probability} (or just \emph{prior}) is introduced that can be interpreted as the \emph{prior belief} or \emph{a priori} knowledge.

\begin{equation} \text{posterior probability} = \frac{\text{conditional probability} \cdot \text{prior probability}}{\text{evidence}} \end{equation}

In the context of pattern classification, the prior probabilities are also called \emph{class priors}, which describe "the general probability of encountering a particular class." In the case of spam classification, the priors could be formulated as 

\begin{equation}P(\text{spam})=\text{"the probability that any new message is a spam message"} \end{equation} and 

\begin{equation} P(\text{ham})= 1-P(\text{spam}). \end{equation}

If the priors are following a uniform distribution, the posterior probabilities will be entirely determined by the class-conditional probabilities and the evidence term. And since the evidence term is a constant, the decision rule will entirely depend on the class-conditional probabilities (similar to a frequentist's approach and maximum-likelihood estimate).

Eventually, the \emph{a priori} knowledge can be obtained, e.g., by consulting a domain expert or by estimation from the training data (assuming that the training data is \emph{i.i.d.} and a representative sample of the entire population. The maximum-likelihood estimate approach can be formulated as

\begin{equation} \hat{P}(\omega_j) = \frac{N_{\omega_j}}{N_c}  \end{equation}

\begin{itemize}
	\item $N_{\omega_j}$: Count of samples from class $\omega_j$.
	\item $N_c$: Count of all samples.
\end{itemize}

And in context of  \emph{spam classification}:

\begin{equation}\hat{P}(\text{spam}) = \frac{\text{\# of spam messages in training data}}{\text{ \# of all messages in training data}} \end{equation}

\begin{figure}[h!]
\includegraphics[width=\linewidth]{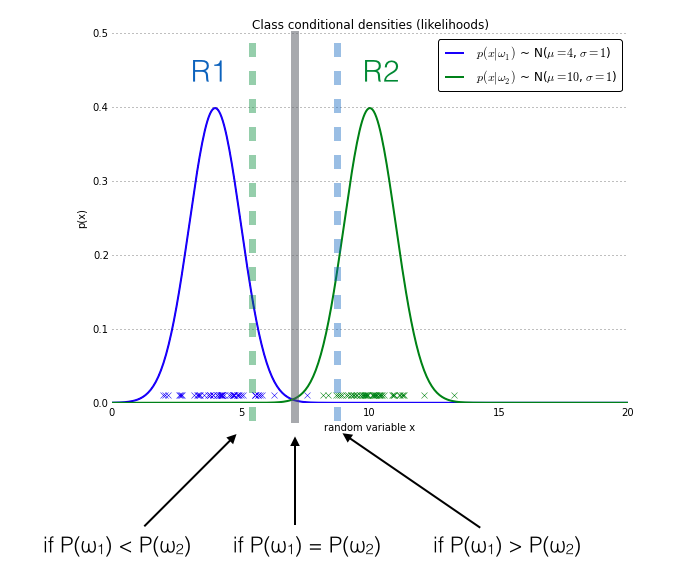}
\caption{The effect of prior probabilities on the decision regions. The figure shows an 1-dimensional random sample from two different classes (blue and green crosses). The data points of both the blue and the green class are normally distributed with standard deviation 1, and the bell curves denote the class-conditional probabilities. If the class priors are equal, the decision boundary of a naive Bayes classifier is placed at the center between both distributions (gray bar). An increase of the prior probability of the blue class ($\omega_1$) leads to an extension of the decision region R1 by moving the decision boundary (blue-dotted bar) towards the other class and vice versa.}
\label{fig:effect_priors}
\end{figure}

Figure \ref{fig:effect_priors} illustrates the effect of the prior probabilities on the decision rule. Given an 1-dimensional pattern \textbf{x} (continuous attribute, plotted as "x" symbols) that follows a normal distribution and belongs to one out of two classes (\emph{blue} and \emph{green}). The patterns from the first class ($\omega_1=\text{blue}$) are drawn from a normal distribution with mean  $x=4$ and a standard deviation $\sigma=1$. The probability distribution of the second class ($\omega_2=\text{green}$) is centered at x=10 with a similar standard deviation of $\sigma=1$. The bell-curves denote the probability densities of the samples that were drawn from the two different normal distributions. Considering only the class conditional probabilities, the maximum-likelihood estimate in this case would be  

\begin{equation} 
\begin{split}
&P(x=4 \mid \omega_1) \approx 0.4 \text{ and } P(x=10 \mid \omega_1) < 0.001\\
&P(x=4 \mid \omega_2) < 0.001 \text{ and } P(x=10 \mid \omega_2) \approx 0.4.
\end{split}
\end{equation}

Now, given uniform priors, that is $P(\omega_1) = P(\omega_2) = 0.5$, the decision rule would be entirely dependent on those class-conditional probabilities, so that the decision rule would fall directly between the two distributions

\begin{equation}  P(x \mid \omega_1) = P(x \mid \omega_2).\end{equation}

However, if the prior probability was $P(\omega_1) > 0.5$, the decision region of class $\omega_1$ would expand as shown in Figure \ref{fig:effect_priors}. In the context of spam classification, this could be interpreted as encountering a new message that only contains words which are equally likely to appear in \emph{spam} or \emph{ham} messages. In this case, the decision would be entirely dependent on \emph{prior knowledge}, e.g., we could assume that a random message is in 9 out of 10 cases not \emph{spam} and therefore classify the new message as \emph{ham}.

\subsection{Evidence}
\label{sec:evidence}

After defining the \emph{class-conditional probability} and \emph{prior probability}, there is only one term missing in order to compute \emph{posterior probability}, that is the \emph{evidence}.

\begin{equation} \text{posterior probability} = \frac{\text{conditional probability} \cdot \text{prior probability}}{\text{evidence}} \end{equation}

The evidence $P(\textbf x)$ can be understood as the probability of encountering a particular pattern $\textbf x$ independent from the class label. Given the more formal definition of posterior probability
\begin{equation} P(\omega_j \mid \textbf x_i) = \frac{P(\textbf x_i \mid \omega_j) \cdot P(\omega_j)}{P(\textbf x_i)}, \end{equation}

the evidence can be calculated as follows ($\omega_j^C$ stands for "complement" and basically translates to "\textbf{not} class $\omega_j$." ):

\begin{equation}P(\textbf x_i) = P(\textbf x_i \mid \omega_j) \cdot P(\omega_j) + P(\textbf x_i \mid \omega_j^C) \cdot P(\omega_j^C)\end{equation}

 Although the evidence term is required to accurately calculate the posterior probabilities, it can be removed from the \emph{decision rule} "Classify sample $\textbf x_i$ as $\omega_1$ if $P(\omega_1 \mid \textbf x_i) > P(\omega_2 \mid \textbf x_i)$ else classify the sample as $\omega_2$,"   since it is merely a scaling factor:

\begin{equation}  \frac{P(\textbf x_i \mid \omega_1) \cdot P(\omega_1)}{P(\textbf x_i)} > \frac{P(\textbf x_i \mid \omega_2) \cdot P(\omega_2)}{P(\textbf x_i)} \end{equation}

\begin{equation}\propto P(\textbf x_i \mid \omega_1) \cdot P(\omega_1) > P(\textbf x_i \mid \omega_2) \cdot P(\omega_2)\end{equation}

\subsection{Multinomial Naive Bayes - A Toy Example}
\label{sec:multinomial_naive_bayes-a_toy_example}

\begin{figure}[h!]
\includegraphics[scale=0.5]{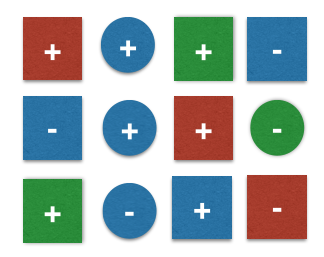}
\caption{A simple toy dataset of 12 samples 2 different classes $+, -$  . Each sample consists of 2 features: color and geometrical shape.}
\label{fig:toy_dataset}
\end{figure}

After covering the basics concepts of a naive Bayes classifier, the \emph{posterior probabilities} and \emph{decision rules}, let us walk through a simple toy example based on the training set shown in Figure \ref{fig:toy_dataset}.

Let

\begin{itemize}
	\item $\omega_j$ be the class labels: $\omega_j \in \{+, -\}$
	\item and $\textbf x_i$  be the 2-dimensional feature vectors: $\textbf x_i = [x_{i1} \; x_{i2}], \quad x_{i1} \in \{ \text{blue}, \text{green}, \text{red}, \text{yellow} \}, \quad x_{i2} \in \{\text{circle}, \text{square} \}.$
\end{itemize}

The 2 class labels are $\omega_j \in \{+, -\}$ and the feature vector for sample $i$ can be written as

\begin{equation}
\begin{split}
& \textbf{x}_i = [x_{i1} \; x_{i2}] \\ 
& \text{for } i \in \{1, 2, ...,  n\}, \; \text{ with } n=12 \\
& \text{and } x_{i1} \in \{ \text{blue}, \text{green}, \text{red}, \text{yellow} \}, \quad x_{i2} \in \{\text{circle}, \text{square} \}
\end{split}
\end{equation}
 
 The task now is to classify a new sample --- pretending that we don't know that its true class label is "+":

 \begin{figure}[h!]
\includegraphics[scale=0.5]{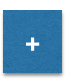}
\caption{A new sample from class $+$ and the features $\textbf{x} = \text{[blue, square]}$ that is to be classified using the training data in Figure \ref{fig:toy_dataset}.}
\label{fig:new_sample1}
\end{figure}
 
\subsubsection{Maximum-Likelihood Estimates}

The \emph{decision rule} can be defined as

\begin{equation} 
\begin{split}
& \text{Classify sample as } +  \text{ if}   \\
& P(\omega=\text{+} \mid \textbf x = \text{[blue, square]}) \geq P(\omega=\text{-} \mid \textbf x = \text{[blue, square]})\\
& \text{else classify sample as} -. 
\end{split}
\end{equation}

Under the assumption that the samples are \emph{i.i.d}, the  \emph{prior probabilities} can be obtained via the maximum-likelihood estimate (i.e., the frequencies of how often each class label is represented in the training dataset):

\begin{equation} 
\begin{split}
& P(\text{+}) = \frac{7}{12} = 0.58 \\
& P(\text{-}) = \frac{5}{12} = 0.42
\end{split}
\end{equation}

Under the \emph{naive} assumption that the features "color" and "shape" are mutually independent, the \emph{class-conditional probabilities} can be calculated as a simple product of the individual conditional probabilities.

Via maximum-likelihood estimate, e.g., $P(\text{blue} \mid -)$ is simply the frequency of observing a "blue" sample among all samples in the training dataset that belong to class $-$.

\begin{equation} 
\begin{split}
& P(\textbf{x} \mid +) = P(\text{blue} \mid +) \cdot P(\text{square} \mid +) = \frac{3}{7} \cdot \frac{5}{7} = 0.31  \\
& P(\textbf{x} \mid -) = P(\text{blue} \mid -) \cdot P(\text{square} \mid -) = \frac{3}{5} \cdot \frac{3}{5} = 0.36 
\end{split}
\end{equation}

Now,  the \emph{posterior probabilities} can be simply calculated as the product of the class-conditional and prior probabilities:

\begin{equation} 
\begin{split}
& P(+ \mid \textbf{x}) = P(\textbf{x} \mid +) \cdot P(+) = 0.31  \cdot  0.58 = 0.18  \\
& P(- \mid \textbf{x}) = P(\textbf{x} \mid -) \cdot P(-) = 0.36  \cdot  0.42 = 0.15
\end{split}
\end{equation}

\subsubsection{Classification}

Putting it all together, the new sample can be classified by plugging in the posterior probabilities into the decision rule:

\begin{equation} 
\begin{split}
&\text{If } P(+ \mid \textbf x) \geq P(\text{-} \mid \textbf{x})  \\
&\text{classify as }+, \\
& \text{else } \text{classify as } -
\end{split}
\end{equation} 

Since $0.18 > 0.15$  the sample can be classified as $+$. Taking a closer look at the calculation of the posterior probabilities, this simple example demonstrates the effect of the prior probabilities affected on the decision rule. If the prior probabilities were equal for both classes, the new pattern would be classified as $-$ instead of $+$. This observation also underlines the importance of \emph{representative} training datasets; in practice, it is usually recommended to additionally consult a domain expert in order to define the prior probabilities.

\subsubsection{Additive Smoothing}
\label{sec:additive_smoothing}

The classification was straight-forward given the sample in Figure \ref{fig:new_sample1}. A trickier case is a sample that has a "new" value for the color attribute that is not present in the training dataset, e.g., \emph{yellow}, as shown in Figure \ref{fig:new_sample1}.

 \begin{figure}[h!]
\includegraphics[scale=0.5]{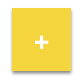}
\caption{A new sample from class $+$ and the features $\textbf{x} = \text{[yellow, square]}$ that is to be classified using the training data in Figure \ref{fig:toy_dataset}.}
\label{fig:new_sample2}
\end{figure}

If the color \emph{yellow} does not appear in our training dataset, the class-conditional probability will be 0, and as a consequence, the posterior probability will also be 0 since the posterior probability is the product of the prior and class-conditional probabilities.

\begin{equation} 
\begin{split}
& P(\omega_1 \mid \textbf x) = 0  \cdot  0.42 = 0 \\
& P(\omega_2 \mid \textbf x) = 0  \cdot  0.58 = 0 
\end{split}
\end{equation} 

In order to avoid the problem of \emph{zero} probabilities, an additional smoothing term can be added to the \emph{multinomial Bayes} model. The most common variants of additive smoothing are the so-called \emph{Lidstone smoothing} ($\alpha<1$) and  \emph{Laplace smoothing} ($\alpha=1$).

\begin{equation} \hat{P}(x_i \mid \omega_j) = \frac{N_{x_i, \omega_j}+\alpha}{N_{\omega_j} + \alpha \, d}  \quad (i = (1, ..., d))\end{equation}

where

\begin{itemize}
\item $N_{x_i, \omega_j}$: Number of times feature $x_i$ appears in samples from class $\omega_j$.
\item  $N_{\omega_j}$: Total count of all features in class $\omega_j$.
\item $\alpha$: Parameter for additive smoothing. 
\item $d$: Dimensionality of the feature vector $\textbf x = [x_1, ..., x_d]$.

\end{itemize}

 \section{Naive Bayes and Text Classification}

This section will introduce some of the main concepts and procedures that are needed to apply the naive Bayes model to text classification tasks. Although the examples are mainly concerning a two-class problem --- classifying text messages as \emph{spam} or \emph{ham} --- the same approaches are applicable to multi-class problems such as classification of documents into different topic areas (e.g., "Computer Science", "Biology", "Statistics", "Economics", "Politics", etc.).

\subsection{The Bag of Words Model}
\label{sec:the_bag_of_words_model}

One of the most important sub-tasks in pattern classification are \emph{feature extraction} and \emph{selection}; the three main criteria of good features are listed below:

\begin{itemize}
	\item \emph{Salient}. The features are important and meaningful with respect to the problem domain.
	\item \emph{Invariant}. Invariance is often described in context of image classification: The features are insusceptible to distortion, scaling, orientation, etc. A nice example is given by C. Yao \emph{et al.} in \emph{Rotation-Invariant Features for Multi-Oriented Text Detection in Natural Images} \cite{yao2013rotation}.
	\item \emph{Discriminatory.} The selected features bear enough information to distinguish well between patterns when used to train the classifier.
\end{itemize}

Prior to fitting the model and using machine learning algorithms for training, we need to think about how to best represent a text document as a feature vector. A commonly used model in \emph{Natural Language Processing} is the so-called \emph{bag of words} model. The idea behind this model really is as simple as it sounds. First comes the creation of the \emph{vocabulary} --- the collection of all different words that occur in the training set and each word is associated with a count of how it occurs. This vocabulary can be understood as a set of non-redundant items where the order doesn't matter. Let $D_1$ and $D_2$ be two documents in a training dataset:

\begin{itemize}
	\item $D_1$: "Each state has its own laws."
	\item $D_2$: "Every country has its own culture."
\end{itemize}

Based on these two documents, the vocabulary could be written as \\

\begin{equation} 
\begin{split}
&V= \{each: 1, state: 1, has: 2, its: 2, own: 2, \\
& laws: 1, every: 1, country: 1, culture: 1\} 
\end{split}
\end{equation}

The vocabulary can then be used to construct the $d$-dimensional feature vectors for the individual documents where the dimensionality is equal to the number of different words in the vocabulary ($d=|V|$). This process is called \emph{vectorization}.

\begin{table}[h]

\caption{Bag of words representation of two sample documents $D_1$ and $D_2$.}
\begin{tabular}{l l l l l l l l l l l} \hline
& each   & state & has & its & own & laws & every & country & culture &   \\ \hline
$\textbf{x}_{D1}$             & 1     & 1   & 1   & 1   & 1    & 1     & 0       & 0       & 0  \\ 
$\textbf{x}_{D2}$             & 0     & 0   & 1   & 1   & 1    & 0     & 1       & 1       & 1 \\ 
$\sum$ & 1     & 1   & 2   & 2   & 2    & 1     & 1       & 1       & 1 \\ \hline
\end{tabular}
\label{fig:bag_of_words}
\end{table}

Given the example in Table \ref{fig:bag_of_words} one question is whether the 1s and 0s of the feature vectors are binary counts (1 if the word occurs in a particular document, 0 otherwise) or absolute counts (how often the word occurs in each document). The answer depends  on which probabilistic model is used for the naive Bayes classifier: The \emph{Multinomial} or \emph{Bernoulli} model --- more on the probabilistic models in Section \ref{sec:bernoulli_bayes} and Section \ref{sec:multinomial_bayes}.

\subsubsection{Tokenization}
\label{sec:tokenization} 

\emph{Tokenization} describes the general process of breaking down a text corpus into individual elements that serve as input for various natural language processing algorithms. Usually, tokenization is accompanied by other optional processing steps, such as the removal of stop words and punctuation characters, stemming or lemmatizing, and the construction of \emph{n-grams}. Below is an example of a simple but typical tokenization step that splits a sentence into individual words, removes punctuation, and converts all letters to lowercase.

\begin{table}[H]
\caption{Example of tokenization.}
\begin{center}
\begin{tabular}{ | c | }

\hline
A swimmer likes swimming, thus he swims. \\ \hline
\end{tabular}

$\downarrow$

\begin{tabular}{ | c | c | c |  c | c | c | c |}
\hline
a & swimmer & likes & swimming & thus & he & swims \\ \hline
\end{tabular}
\end{center}

\end{table}

\subsubsection{Stop Words}
\label{sec:stopwords}

\emph{Stop words} are words that are particularly common in a text corpus and thus considered as rather un-informative (e.g., words such as \emph{so}, \emph{and}, \emph{or}, \emph{the}, ..."). One approach to stop word removal is to search against a language-specific stop word dictionary. An alternative approach is to create a  \emph{stop list} by sorting all words in the entire text corpus by frequency. The stop list --- after conversion into a \emph{set} of non-redundant words --- is then used to remove all those words from the input documents that are ranked among the top \emph{n} words in this stop list.

\begin{table}[H]
\caption{Example of stop word removal.}
\begin{center}
\begin{tabular}{ | c | }
\hline
A swimmer likes swimming, thus he swims. \\ \hline
\end{tabular}

$\downarrow$

\begin{tabular}{ | c | c | c |  c | c | c | }
\hline
swimmer & likes & swimming & , &  swims &. \\ \hline
\end{tabular}
\end{center}
\end{table}

\subsubsection{Stemming and Lemmatization} 
\label{sec:stemming_and_lemmatization}

\emph{Stemming} describes the process of transforming a word into its root form. The original stemming algorithm was developed my Martin F. Porter in 1979 and is hence known as \emph{Porter stemmer} \cite{porter1980algorithm}. 

\begin{table}[H]
\caption{Example of Porter stemming.}
\begin{center}
\begin{tabular}{ | c | }
\hline
A swimmer likes swimming, thus he swims. \\ \hline
\end{tabular}

$\downarrow$

\begin{tabular}{ | c | c | c |  c | c | c | c | c | c | }
\hline
a & swimmer &  like & swim & , & thu & he & swim & .\\ \hline
\end{tabular}
\end{center}
\end{table}

Stemming can create non-real words, such as "thu" in the example above. In contrast to stemming, \emph{lemmatization} aims to obtain the canonical (grammatically correct) forms of the words, the so-called \emph{lemmas}. Lemmatization is computationally more difficult and expensive than stemming, and in practice, both stemming and lemmatization have little impact on the performance of text classification \cite{toman2006influence}.

\begin{table}[H]
\caption{Example of lemmatization.}
\begin{center}
\begin{tabular}{ | c | }
\hline
A swimmer likes swimming, thus he swims. \\ \hline
\end{tabular}

$\downarrow$

\begin{tabular}{ | c | c | c |  c | c | c | c | c | c | }
\hline
A & swimmer &  like & swimming & , & thus & he & swim & .\\ \hline
\end{tabular}
\end{center}
\end{table}

The stemming and lemmatization examples were created by using the Python NLTK library (\href{http://www.nltk.org}{http://www.nltk.org}).

\subsubsection{\emph{N}-grams}
\label{sec:n-grams} 

In the \emph{n-gram} model, a token can be defined as a sequence of \emph{n} items. The simplest case is the so-called \emph{unigram} (1-gram) where each word consists of exactly one word, letter, or symbol. All previous examples were unigrams so far. Choosing the optimal number \emph{n} depends on the language as well as the particular application. For example, Andelka Zecevic found in his study that \emph{n}-grams with $3 \le n \le 7$ were the best choice to determine authorship of Serbian text documents \cite{zevcevic2011n}. In a different study, the \emph{n}-grams of size $4 \le n \le 8$ yielded the highest accuracy in authorship determination of English text books \cite{kevselj2003n} and Kanaris \emph{e. al.} report that \emph{n}-grams of size 3 and 4 yield good performances in anti-spam filtering of e-mail messages  \cite{kanaris2007words}.

\begin{itemize}
\item unigram (1-gram):   

\begin{tabular}{ | c | c | c |  c | c | c | c | c | c | }
\hline
a & swimmer &  likes & swimming & thus & he & swims\\ \hline
\end{tabular}

\item bigram (2-gram):  

\begin{tabular}{ | c | c | c |  c | c | }
\hline
a swimmer & swimmer likes &  likes swimming & swimming thus & ... \\ \hline
\end{tabular}

\item trigram (3-gram):  

\begin{tabular}{ | c | c | c |  c | c | }
\hline
a swimmer likes & swimmer likes swimming &  likes swimming thus &  ... \\ \hline
\end{tabular}

\end{itemize}

\subsection{The Decision Rule for Spam Classification}
\label{sec:decision_rule_spam}

In context of spam classification the decision rule of a naive Bayes classifier based on the posterior probabilities can be expressed as

\begin{equation}
\begin{split} 
    & \text{if } P(  \omega = \text{spam} \mid \textbf{x}) \ge P(\omega = \text{ham} \mid \textbf{x})  \text{ classify as spam, }\\
    &  \text{else classify as ham. }
\end{split}
\end{equation}

As described in Section \ref{sec:posterior_probabilities_1} the posterior probability is the product of the class-conditional probability and the prior probability; the evidence term in the denominator can be dropped since it is constant for both classes.

\begin{equation}
\begin{split} 
    &  P(\omega = \text{spam} \mid \textbf{x}) = P(\textbf{x} \mid \omega = \text{spam}) \cdot P(\text{spam}) \\
    &  P(\omega = \text{ham} \mid \textbf{x}) = P(\textbf{x} \mid \omega = \text{ham}) \cdot P(\text{ham})
\end{split}
\end{equation}

The prior probabilities can be obtained via the maximum-likelihood estimate based on the frequencies of spam and ham messages in the training dataset:

\begin{equation}
\begin{split} 
  & \hat{P}(\omega = \text{spam}) = \frac {\text{\# of spam msg.}}{\text{\# of all msg.}} \\
   &  \hat{P}(\omega = \text{ham}) = \frac {\text{\# of ham msg.}}{\text{\# of all msg.}}
\end{split}
\end{equation}

Assuming that the words in every document are conditionally independent (according to the \emph{naive} assumption), two different models can be used to compute the class-conditional probabilities: The \emph{Multi-variate Bernoulli} model (Section \ref{sec:bernoulli_bayes}) and the \emph{Multinomial} model (Section \ref{sec:multinomial_bayes}).

\subsection{Multi-variate Bernoulli Naive Bayes}
\label{sec:bernoulli_bayes}

The \emph{Multi-variate Bernoulli} model is based on binary data: Every token in the feature vector of a document is associated with the value 1 or 0. The  feature vector has $m$ dimensions where $m$ is the number of words in the whole vocabulary (in Section \ref{sec:the_bag_of_words_model}); the value 1 means that the word occurs in the particular document, and 0 means that the word does
not occur in this document. The Bernoulli trials can be written as

\begin{equation} P(\textbf x \mid \omega_j) = \prod_{i=1}^{m} P(x_i \mid \omega_j)^b \cdot (1-P(x_i \mid  \omega_j))^{(1-b)}  \quad (b \in {0,1}).
\end{equation}

Let  $\hat{P}(x_i \mid \omega_j)$ be the maximum-likelihood estimate that a particular word (or token) $x_i$ occurs in class $\omega_j$. 

\begin{equation} \hat{P}(x_i \mid \omega_j) = \frac{df_{xi, y} + 1}{df_y + 2} \end{equation} 

where

\begin{itemize}
\item $df_{xi, y}$ is the number of documents in the training dataset that contain the feature $x_i$ and belong to class $\omega_j$.
\item  $df_y$ is the number of documents in the training dataset that belong to class $\omega_j$.
\item  +1 and +2 are the parameters of \emph{Laplace smoothing} (Section \ref{sec:additive_smoothing}).
\end{itemize}

\subsection{Multinomial Naive Bayes}
\label{sec:multinomial_bayes}

\subsubsection{Term Frequency}
\label{sec:term_frequency}

A alternative approach to characterize text documents --- rather than binary values ---  is the \emph{term frequency (tf(t, d))}. The term frequency is typically defined as the number of times a given term \emph{t} (i.e., word or token) appears in a document \emph{d} (this approach is sometimes also called \emph{raw frequency}). In practice, the term frequency is often normalized by dividing the raw term frequency by the document length.

\begin{equation}\text{normalized term frequency} = \frac{tf(t, d)}{n_d}\end{equation}

where 

\begin{itemize}
\item  $tf(t, d)$: Raw term frequency (the count of term $t$ in document $d$).
\item $n_d$: The total number of terms in document $d$.
\end{itemize}

The term frequencies can then be used to compute the maximum-likelihood estimate based on the training data to estimate the class-conditional probabilities in the multinomial model:

\begin{equation} \hat{P}(x_i \mid \omega_j) = \frac{\sum tf(x_i, d \in \omega_j) + \alpha}{\sum N_{d \in \omega j} + \alpha \cdot V} \end{equation} 

where

\begin{itemize}
\item  $x_i$: A word from the feature vector $\textbf x$ of a particular sample.
\item  $\sum tf(x_i, d \in \omega_j)$: The sum of raw term frequencies of word $x_i$ from all documents in the training sample that belong to class $\omega_j$.
\item  $\sum N_{d \in \omega j}$: The sum of all term frequencies in the training dataset for class $\omega_j$.
\item $\alpha$: An additive smoothing parameter ($\alpha = 1$ for Laplace smoothing).
\item  $V$: The size of the vocabulary (number of different words in the training set).
\end{itemize}

The class-conditional probability of encountering the text $\textbf x$ can be calculated as the product from the likelihoods of the individual words (under the \emph{naive} assumption of conditional independence).

\begin{equation} P(\textbf x \mid \omega_j) = P(x_1 \mid \omega_j) \cdot P(x_2 \mid \omega_j) \cdot   \dotsc \cdot P(x_n \mid \omega_j) = \prod_{i= 1}^{m}  P(x_i \mid \omega_j)  \end{equation}

\subsubsection{Term Frequency - Inverse Document Frequency (Tf-idf)}
\label{sec:tf-idf}

The \emph{term frequency - inverse document frequency (Tf-idf)} is another alternative for characterizing text documents. It can be understood as a weighted \emph{term frequency}, which is especially useful if stop words have  not been removed from the text corpus. The Tf-idf approach assumes that the importance of a word is inversely proportional to how often it occurs across all documents. Although Tf-idf is most commonly used to rank documents by relevance in different text mining tasks, such as page ranking by search engines, it can also be applied to text classification via naive Bayes.

\begin{equation}\text{Tf-idf} = tf_n(t,d) \cdot idf(t)\end{equation}

Let $tf_n(d,f)$ be the normalized term frequency, and $idf$, the inverse document frequency, which can be calculated as follows

\begin{equation}idf(t) = \log\Bigg(\frac{n_d}{n_d(t)}\Bigg), \end{equation}

where

\begin{itemize}
\item  $n_d$: The total number of documents.
\item  $n_d(t)$: The number of documents that contain the term $t$.
\end{itemize}

\subsubsection{Performances of the Multi-variate Bernoulli and Multinomial Model}

Empirical comparisons provide evidence that the multinomial model tends to outperform the multi-variate Bernoulli model if the vocabulary size is relatively large \cite{mccallum1998comparison}. However, the performance of machine learning algorithms is highly dependent on the appropriate choice of features. In the case of naive Bayes classifiers and text classification, large differences in performance can be attributed to the choices of  stop word removal, stemming, and token-length \cite{rudner2002automated}. In practice, it is recommended that the choice between a multi-variate Bernoulli or multinomial model for text classification should precede comparative studies including different combinations of feature extraction and selection steps.

\section{Variants of the Naive Bayes Model}
\label{sec:naive_bayes_variants}

So far, we have seen two different models for categorical data, namely, the multi-variate Bernoulli (Section \ref{sec:bernoulli_bayes}) and multinomial (Section \ref{sec:multinomial_bayes}) models --- and two different approaches for the estimation of class-conditional probabilities. In Section \ref{sec:continuous_variables}, we will take a brief look at a third model: \emph{Gaussian naive Bayes}.

\subsection{Continuous Variables}
\label{sec:continuous_variables}

Text classification is a typical case of categorical data, however, naive Bayes can also be used on continuous data. The \emph{Iris} flower data set would be a simple example for a supervised classification task with continuous features: The Iris dataset contains widths and lengths of petals and sepals measured in centimeters.  One strategy for dealing with continuous data in naive Bayes classification would be to discretize the features and form distinct categories or to use a Gaussian kernel to calculate the class-conditional probabilities. Under the assumption that the probability distributions of the features  follow a normal  (Gaussian)  distribution, the Gaussian naive Bayes model can be written as follows

\begin{equation} P(x_{ik} \mid \omega) = \frac{1}{\sqrt{2\pi\sigma^2_{\omega}}} \exp\left(-\frac{(x_{ik} - \mu_{\omega})^2}{2\sigma^2_{\omega}}\right), \end{equation}

where $\mu$ (the sample mean) and $\sigma$ (the standard deviation) are the parameters that are to be estimated from the training data. Under the naive Bayes assumption of conditional independence, the class-conditional probability can than be computed as the product of the individual probabilities:

\begin{equation} P(\textbf x_i \mid \omega) = \prod_{k=1}^{d} P(\textbf x_{ik} \mid \omega) \end{equation}

\subsection{Eager and Lazy Learning Algorithms}
\label{sec:eager_and_lazy}

Being an \emph{eager learner}, naive Bayes classifiers are known to be relatively fast in classifying new instances. Eager learners are learning algorithms that learn a model from a training dataset as soon as the data becomes available. Once the model is learned, the training data does not have to be re-evaluated in order to make a new prediction. In case of eager learners, the computationally most expensive step is the model building step whereas the classification of new instances is relatively fast.  

\emph{Lazy learners}, however, memorize and re-evaluate the training dataset for predicting the class label of new instances. The advantage of \emph{lazy learning} is that the model building (training) phase is relatively fast. On the other hand, the actual prediction is typically slower compared to eager learners due to the re-evaluation of the training data. Another disadvantage of lazy learners is that the training data has to be retained, which can also be expensive in terms of storage space. A typical example of a lazy learner would be a \emph{k-nearest neighbor} algorithm: Every time a new instance is encountered, the algorithm would evaluate the \emph{k}-nearest neighbors in order to decide upon a class label for the new instance, e.g., via the \emph{majority rule} (i.e., the assignment of the class label that occurs most frequently amongst the \emph{k}-nearest neighbors).

\bibliography{../references/bayes}{}
\bibliographystyle{unsrt}

\end{document}